\relax
\documentclass[letterpaper]{article} % DO NOT CHANGE THIS
\usepackage{ner}  % DO NOT CHANGE THIS
\usepackage{times}  % DO NOT CHANGE THIS
\usepackage{helvet} % DO NOT CHANGE THIS
\usepackage{courier}  % DO NOT CHANGE THIS
\usepackage{graphicx} % DO NOT CHANGE THIS
\usepackage{graphicx}  % DO NOT CHANGE THIS
\usepackage{latexsym}
\usepackage{url}
\usepackage{booktabs}
\usepackage{subfigure}
\usepackage{float}
\usepackage{caption}
\usepackage{amsmath}
\usepackage{multirow}
\usepackage{color}
\usepackage{amsfonts}
\usepackage{xcolor,stfloats}
\usepackage{array}
\usepackage{setspace}
\usepackage{url}
\usepackage{graphicx}  %Required
\aclfinalcopy
\title{Modeling Named Entity Embedding Distribution into Hypersphere}
\author{
	Zhuosheng Zhang$^{1,2,3,}$\thanks{$\ $ These authors contribute equally. $\dagger$ Corresponding author. This Work was done while Bingjie Tang was an intern at SJTU. This work was partially supported by National Key Research and Development Program of China (No. 2017YFB0304100) and Key Projects of National Natural Science Foundation of China (U1836222 and 61733011).}, 
	Bingjie Tang$^{1,2,3,4,*}$,
	Zuchao Li$^{1,2,3}$,
	Hai Zhao$^{1,2,3,\dagger}$
	\\
	$^1$Department of Computer Science and Engineering, Shanghai Jiao Tong University\\
	$^2$Key Laboratory of Shanghai Education Commission for Intelligent Interaction\\
	and Cognitive Engineering, Shanghai Jiao Tong University, Shanghai, China\\
	$^3$MoE Key Lab of Artificial Intelligence, AI Institute, Shanghai Jiao Tong University, Shanghai, China\\
	$^4$Computer Science Department, Brown University, Rhode Island, USA\\
	{\tt \{zhangzs,charlee\}@sjtu.edu.cn, bingjie\_tang@brown.edu, zhaohai@cs.sjtu.edu.cn}
}
 
\begin{document}

\maketitle

\begin{abstract}
		This work models named entity distribution from a way of visualizing topological structure of embedding space, so that we make an assumption that most, if not all, named entities (NEs) for a language tend to aggregate together to be accommodated by a specific hypersphere in embedding space. Thus we present a novel open definition for NE which alleviates the obvious drawback in previous closed NE definition with a limited NE dictionary. Then, we show two applications with introducing the proposed named entity hypersphere model. First, using a generative adversarial neural network to learn a transformation matrix of two embedding spaces, which results in a convenient determination of named entity distribution in the target language, indicating the potential of fast named entity discovery only using isomorphic relation between embedding spaces. Second, the named entity hypersphere model is directly integrated with various named entity recognition models over sentences to achieve state-of-the-art results. Only assuming that embeddings are available, we show a prior knowledge free approach on effective named entity distribution depiction.
\end{abstract}

\section{Introduction}
	
	Named Entity (NE) recognition task intends to identify words or phrases that contain the names of persons, organizations, locations and other NEs alike \cite{carreras2003named,voyer2010hybrid,nothman2013learning,sienvcnik2015adapting,wang2017named}. Accurate NE recognition is generally context dependent, thus the existence of a large amount of annotated data is indispensable for the success of NE recognition. Nevertheless, data annotation may not be always available. For the lack of training data, NE recognition task may seek a cross-lingual solution such as transliteration \cite{tsai2016cross}.
	
	Actually, the NE issue is more complicated than what researchers expect and it is not only a data annotation problem either. For example, the third planet is called \emph{Earth} but not \emph{Venus} due to a common agreement based on a sense of knowledge from real world but not linguistics motivation. NEs are generally related to common sense knowledge part inside human language, which have shown to be very hard defined with a known NE dictionary even for monolingual processing, as new NEs keep emerging always. This means that there will never an NE dictionary that can stably, sufficiently represent NE set for a language and all current NE dictionaries have to be frequently maintained. This work presents a new solution for open NE definition by generally exploring the geometric or topological distribution of NEs for a specific language embedding space and expects to find an effective way mapping one NE distribution from one language to another. In detail, based on the presented visualization of NE distributions in multilingual word embeddings, we summarize a hypersphere model for geometric depiction of NE distribution. By using a transformation matrix learned by generative adversarial network between two embedding spaces, NE hypersphere can be mapped between two languages. Besides considering context independent NE extraction and mapping in embedding spaces, we also show that context dependent NE recognition on sentence can benefit from the proposed NE hypersphere model through presenting its helpful cue.
	
	\begin{figure}
		\centering
		\begin{tabular}{ccc}
			\includegraphics[width=0.27\linewidth]{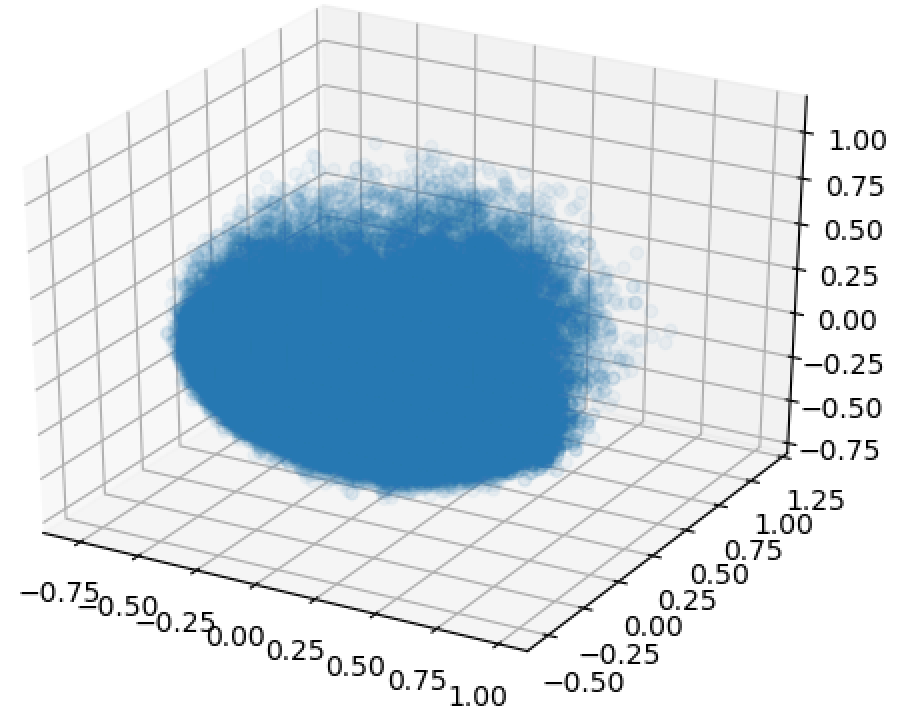}  & 
			\includegraphics[width=0.27\linewidth]{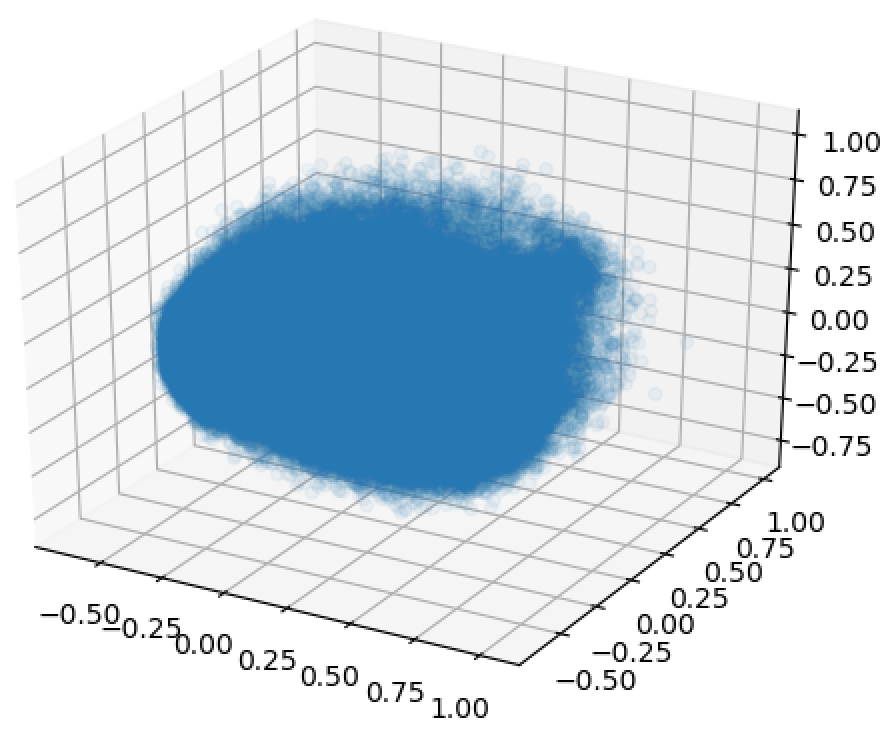}  & 
			\includegraphics[width=0.27\linewidth]{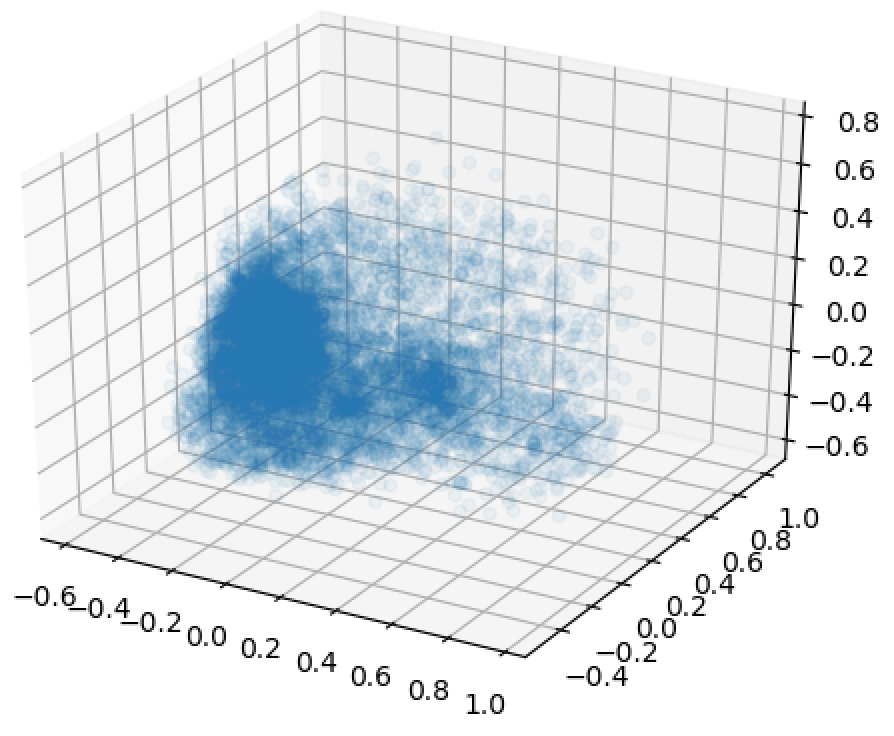} \\ 
			(a) & (b) & (c) \\
		\end{tabular}
		\caption{Distributions of English NE types, (a) person, (b) location, (c) organization.}
		\label{Fig:2}
		\vspace{-0.5em}
	\end{figure}
	
	\begin{figure}
		\centering
		\begin{tabular}{ccc}
			\includegraphics[width=0.3\linewidth]{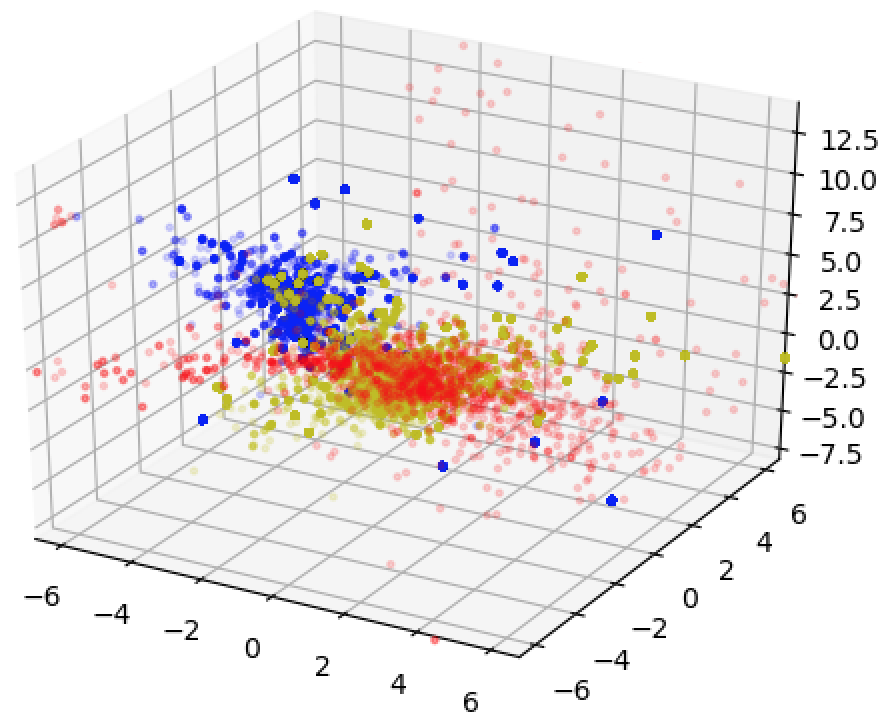}  & 
			\includegraphics[width=0.29\linewidth]{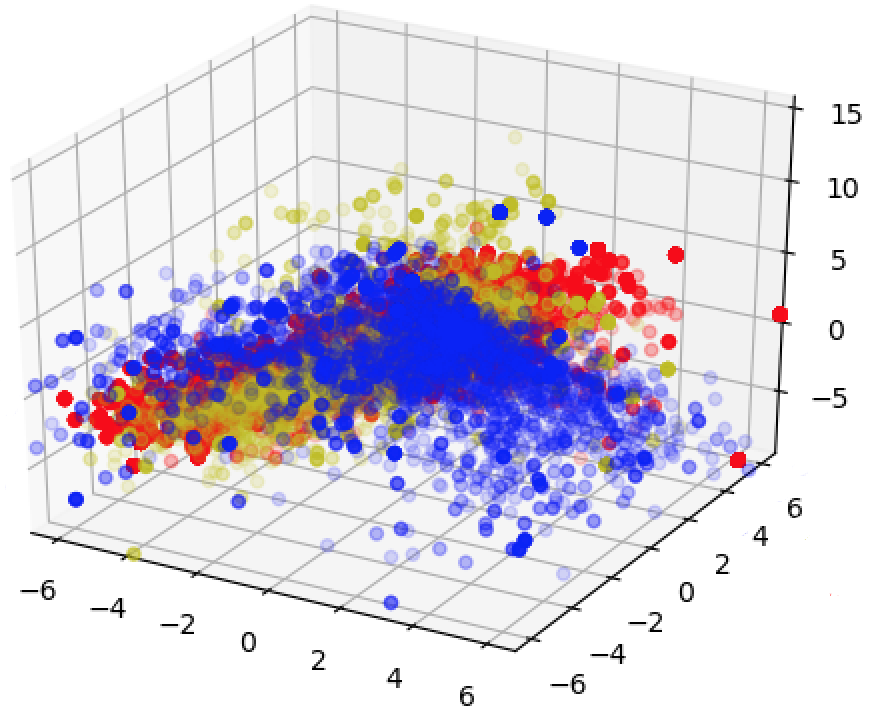}  & 
			\includegraphics[width=0.29\linewidth]{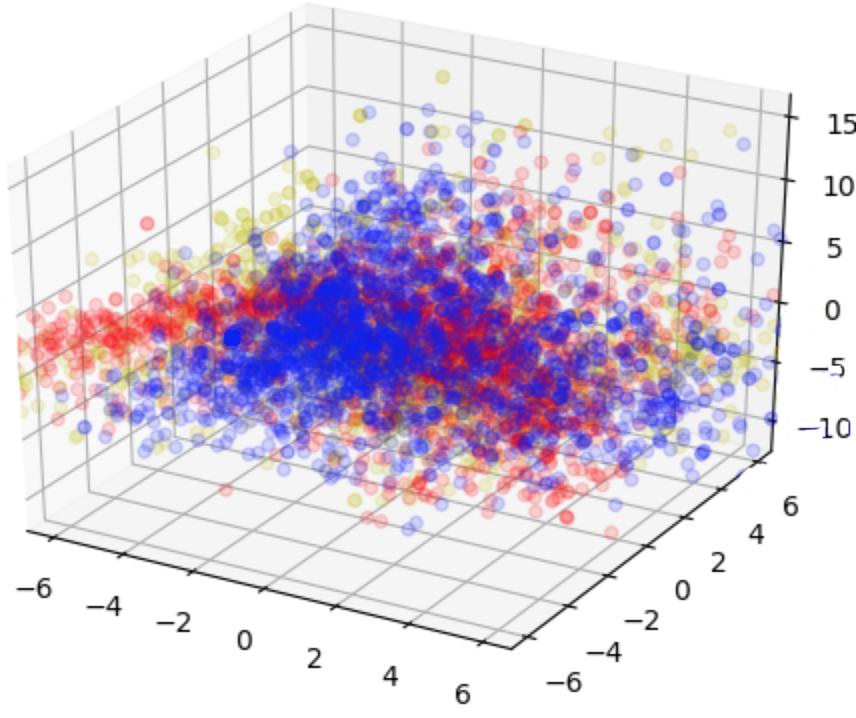} \\ 
			(a) & (b) & (c) \\
		\end{tabular}
		\caption{Distributions of NEs in (a) German, (b) Dutch, and (c) Spanish. Red for person, yellow for locations and blue for organizations.}
		\label{fig-NE-visual}
		\vspace{-0.5em}
	\end{figure}
	The main contributions of this paper are of three aspects. First, we present and analyze the visualization of NE distributions and find a common pattern over NE distribution on multilingual embedding space, thus propose a novel open definition for NEs in any language. Second, utilizing a transformation matrix according to isomorphic relation between two embedding spaces, NE distributions from one language to another can be conveniently and accurately obtained. Third, we show that context dependent named entity recognition from sentence can be further improved from the proposed generally depicted hypersphere model.
	
	\section{Visualization of NE Embeddings}
	
	\begin{table*}
		\begin{minipage}[b]{.5\linewidth}
			\centering
			\begin{tabular}{lcccc}
				\toprule  
				Language& Per& Loc& Org& Total\\
				\midrule  
				English& 80.0K& 98.0K& 8.81K& 187K\\
				Chinese& 22.3K& 169K& 10.8K& 202K\\
				Dutch& 29.0K& 4.56K& 4.17K& 37.7K\\
				German& 1.31K& 76.2K& 17.1K& 94.6K\\
				Spanish& 1.81K& 1.37K& 2.30K& 5.48K\\
				\bottomrule 
			\end{tabular}
			\caption{ Statistics of NE dictionaries }
			\label{table-sts-NE-dictionaries}
			\centering
		\end{minipage}%
		\begin{minipage}[b]{.5\linewidth}
			\centering
			\begin{tabular}{lccc}
				\toprule  
				Language& Corpus Size& Vocab Size&\\
				\midrule  
				English& 14.13GB& 2231K&\\
				Chinese& 1.44GB& 795K&\\
				Dutch& 1GB& 50K&\\
				German& 1GB& 50K&\\
				Spanish& 1GB& 50K&\\
				\bottomrule 
			\end{tabular}
			\caption{ Statistics of Wikipedia Corpus }
			\label{corpus-info}
		\end{minipage}
	\end{table*}
	
	\label{sec-visualization}
	For NE visualization, we use pre-trained word embeddings with the continuous skip-gram model \cite{mikolov2013distributed}. Due to the fact that syntactically and semantically related words tend to appear in similar contexts, this objective of embedding learning is supposed to output similar (i.e., geometrically neighbored) embedding vectors for the related words \cite{seok2016named}. 
	
	This work considers three basic NE types, i.e., person, location and organization names\footnote{Though there are various types of NEs, these three NE types are especially well studied as they are non-trivial for recognition and annotated datasets are commonly available, which thus is our focus in this paper.}. Limited sized self-collected NE dictionaries in Chinese, English and other three languages are used with statistics shown in Table \ref{table-sts-NE-dictionaries}.

	\begin{figure}
		\centering
		\begin{tabular}{ccc}
			\includegraphics[width=0.27\linewidth]{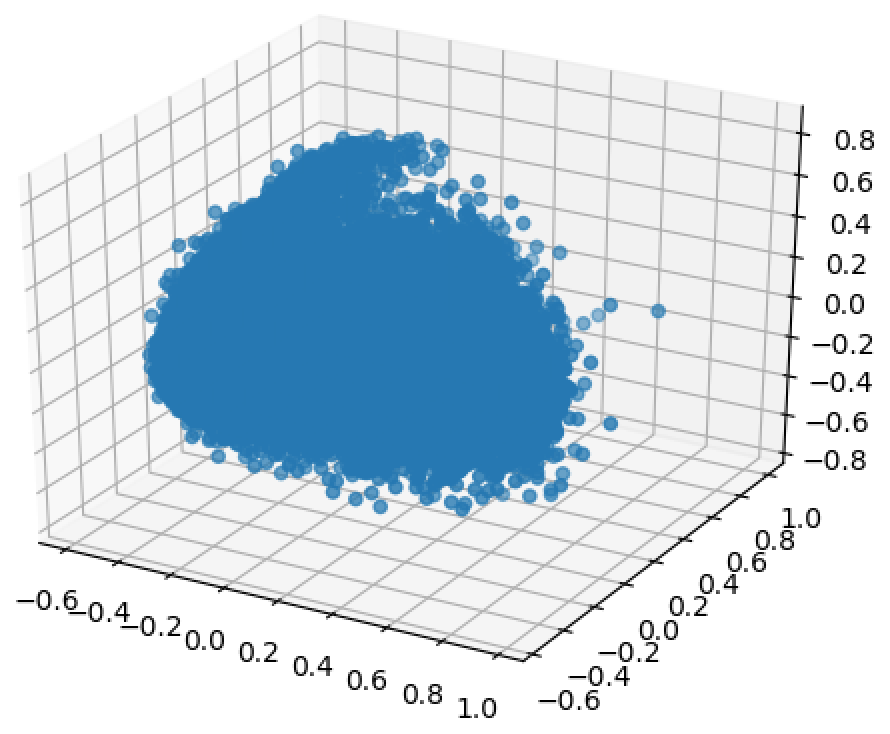}  & 
			\includegraphics[width=0.27\linewidth]{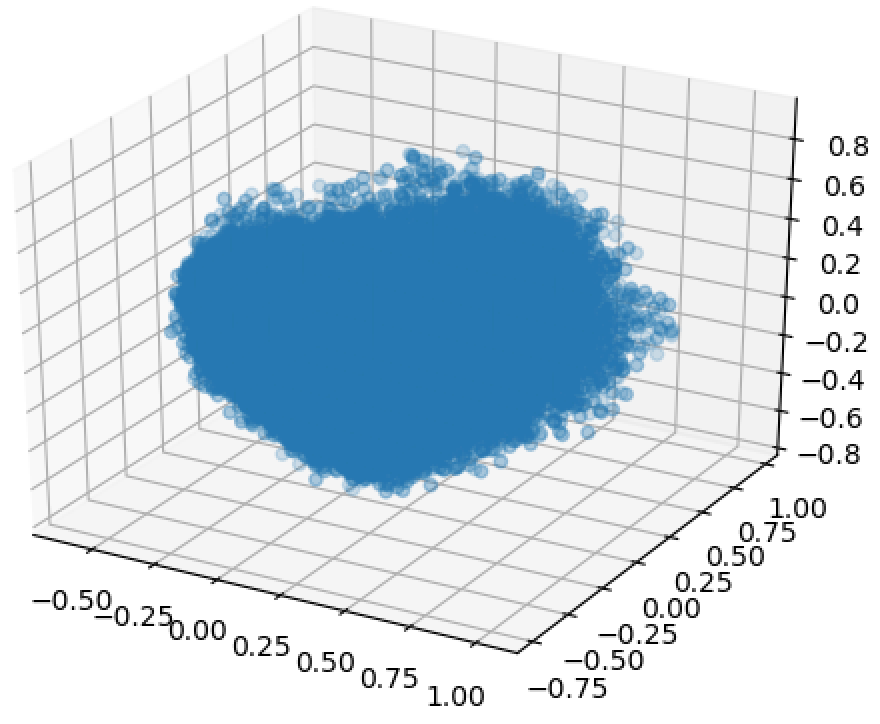}  & 
			\includegraphics[width=0.27\linewidth]{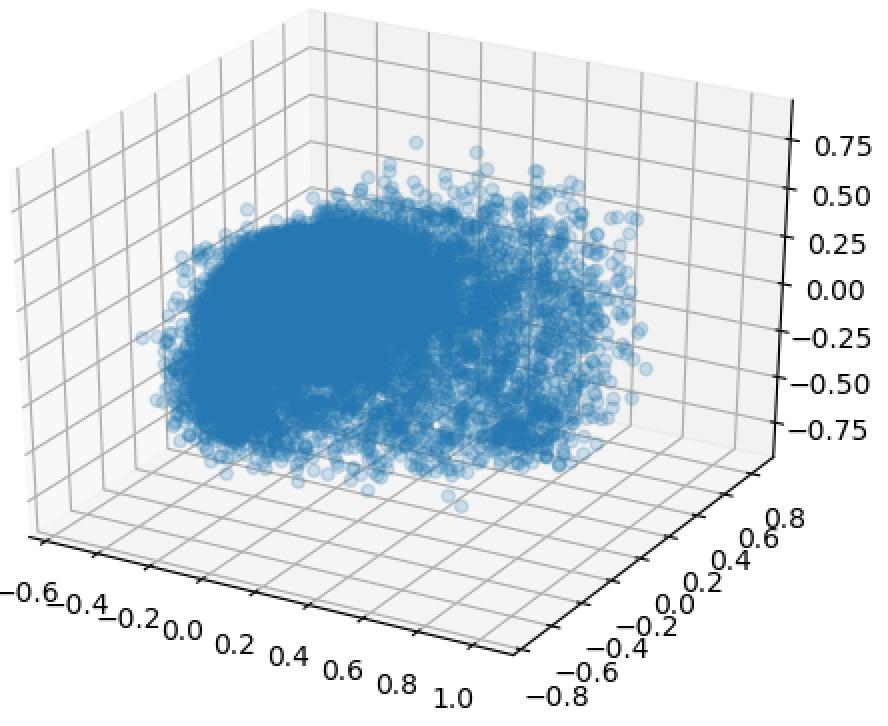} \\ 
			(a) & (b) & (c) \\
		\end{tabular}
		\caption{Distributions of Chinese NE types, (a) person, (b) location, (c) organization.}
		\label{Fig:3}
		\vspace{-0.5em}
	\end{figure}

	\begin{figure}
		\centering
		\begin{tabular}{cc}
			\includegraphics[width=0.48\linewidth]{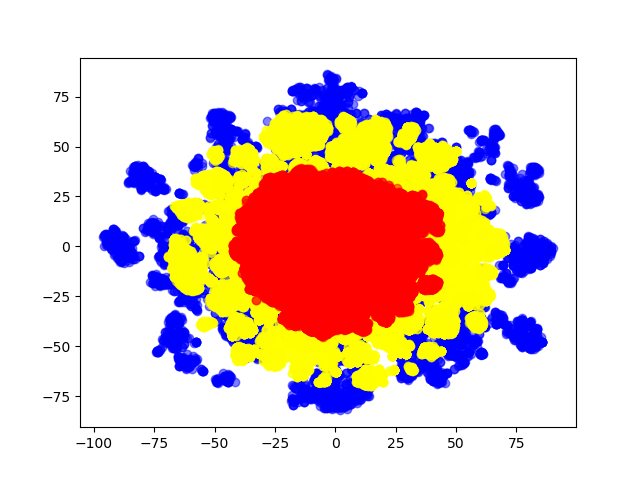} & 
			\includegraphics[width=0.48\linewidth]{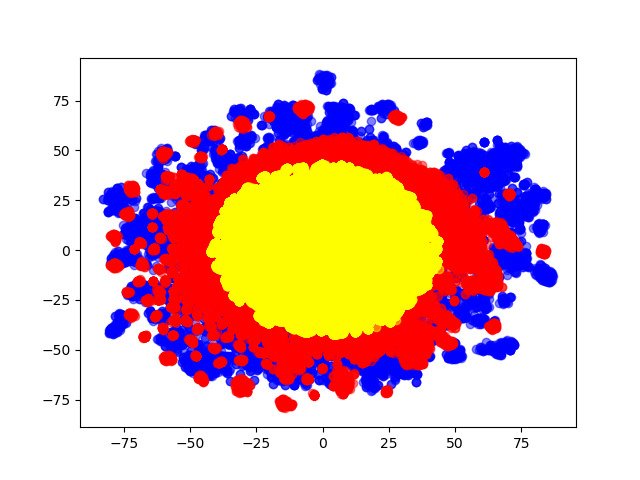} \\
			(a) & (b) \\
		\end{tabular}
		\caption{Distributions of three types of NEs in (a) English, (b) Chinese. Red for person, yellow for locations and blue for organizations.}
		\label{Fig:4}
		\centering
	\end{figure}
	
	Besides the English and Chinese NE visualization, we also draw NE distributions of NEs from German, Dutch and Spanish in Figure \ref{fig-NE-visual}, by using the data from CoNLL 2002 and CoNLL 2003 shared task. Compared to the results in Figures \ref{Fig:2}-\ref{Fig:4} on English and Chinese, the shape of NE distribution is less sphere-like due to the insufficiency of the given NE dictionaries, as they are much smaller than those for Chinese and English. 
	
	Also, we observe that the visualization of the distribution of NEs in German is sparse, even with a larger NE dictionary than Dutch and Spanish, which may be due to relatively more poor quality of German NE dictionary and there is more serious mismatch between the NE dictionary and the corresponding pre-trained embedding. 
	
	As the related NE visualization has been shown far from a proper hypersphere shape, we make a manual verification on the true NE identity by sampling and checking embedding inside the NE hypersphere of German, Dutch and Spanish, but outside our self-collected NEs dictionaries. We found nearly all checked embeddings are truly NE but not in our dictionaries which means that our NE dictionaries over these language are extremely insufficient. Considering the concerned NE dictionaries are not sufficient enough to support a meaningful evaluation, we later dropped out the respective experiments based on them and only focus on English and Chinese.

	We illustrate scattering points of every NEs in embeddings spaces of these languages. For multi-words NEs, we use a simple strategy to represent them with average vector of all member word vectors inside the corresponding NE \footnote{We have tried more advanced strategies including pre-training multi-word embeddings but witnessed unconspicuous performance change thus we  calculate the average vectors to keep simplicity.}. Figures \ref{Fig:2} and \ref{Fig:3} show the distribution of three types of NEs for English and Chinese in 3-D, respectively. Figure \ref{Fig:4} depicts all types of NEs together. To draw the figures, we use pre-trained word embedding on Wikipedia corpus\footnote{https://dumps.wikimedia.org/} in 300-D with Word2Vec skip-gram model \cite{mikolov2013distributed}. Statistics of the Wikipedia corpus is in Table \ref{corpus-info}. Dimensional reduction to 2-D or 3-D is completed by using t-distribution stochastic neighbor embedding (t-SNE) \cite{maaten2008visualizing}.
	
	From the visualization results for English and Chinese, we observe that these NE embeddings are highly concentrated and form a sphere-like shape. 
	Figure \ref{Fig:4} surprisingly shows that besides the gathering feature in 2-D, distributions of all three types of NEs tend to share the same sphere center. 
	
	Considering the shared distribution feature in both English and Chinese, it is promising to identify NEs on Chinese embeddings by properly transforming the identification results of English embeddings.

	\section{NE Hypersphere}
	
	\subsection{Monolingual Hypersphere Model}
	\label{sec-mono-hyper}
	From the visualization of NEs in Figure \ref{fig-NE-visual}, we observe that NEs gather together in different word embedding spaces and show obvious separations among different NE types. 
	
	To accommodate the aggregating NEs by a closure surface with the least defining parameters, we choose the hypersphere\footnote{It should be known that hypersphere will back off to a circle in 2-D space and a sphere in 3-D space.}, which is only defined by two parameters, a center vector $C$ and radius $R$. Thus, we formally define an NE hypersphere model as follows. 
	
	1) For any NE $W$ in a language in terms of word embedding representation, they are subject to,
	$Dist(W,C) < R$,
	where $Dist( , )$ is a predifined distance. Namely, we suppose all known NEs are inside this hypersphere.
	
	2) How likely a word $W$ is to be a true NE can be defined by the distance value $Dist$($W$, $C$). The smaller this value is, the more likely the word $W$ is a true NE. Besides, the hypersphere radius $R$ can be regarded as threshold to distinguish NE and non-NE words in embedding space.
	
	In this paper, we choose $Euclidean$ $distance$ according to our preliminary empirical results\footnote{We tried four different distance measures, including Cosine similarity, Euclidean distance, Earth mover's distance and BM25, Euclidean shows the best.}.
	
	\begin{align}
	ED(W,C) = \sqrt{\sum_{k=1}^{n}(w_{k}-c_{k})^{2}}
	\label{ED}
	\end{align}
	
	Given a known NE dictionary, we evaluate the model by counting the number of NEs from the dictionary which is included in the proposed hypersphere to let the NE hypersphere play as an NE detector. In detail, we define three sets:
	
	T = true named entities, i.e., named entities in dictionary; 
	
	P = predicted named entities, i.e., named entities in hypersphere;
	
	G = T intersects P.
	
	The precision is defined as $\frac{G}{P}$ and the recall is defined as $\frac{G}{T}$. F1-score is then computed from the harmonic average of recall and precision.
	
	All the evaluated data are from NE dictionaries shown in Table \ref{table-sts-NE-dictionaries}, which are supposed to be sufficient and accurate, though not really so. The purpose of the evaluation in monolingual case is just to show to what extent a hypershpere model can accurately depict all NE distribution given by a good enough dictionary.
	
	Note that if we trust NE hypersphere has been an accurate depiction for NE distribution, then we have actually presented an open representation for all NEs in a language. As in theory, there are infinite NEs inside the hypersphere and NE hypersphere contains all known and unknown NEs with only two parameter setting. Furthermore, the NE property of any word/phrase can be simply judged by checking if it is inside the hypersphere, which  is independent of any limited, maybe quite insufficient NE dictionary.
	
	\subsection{Mapping Embedding Spaces}
	
	Based on the assumption that NEs in different languages all fit with the NE hypersphere model, cross-lingual NE detection task can be completed as finding an efficient transformation method between two hyperspheres in different language word embeddings. 
	
\begin{figure}
		\centering
		\includegraphics[width=1\linewidth]{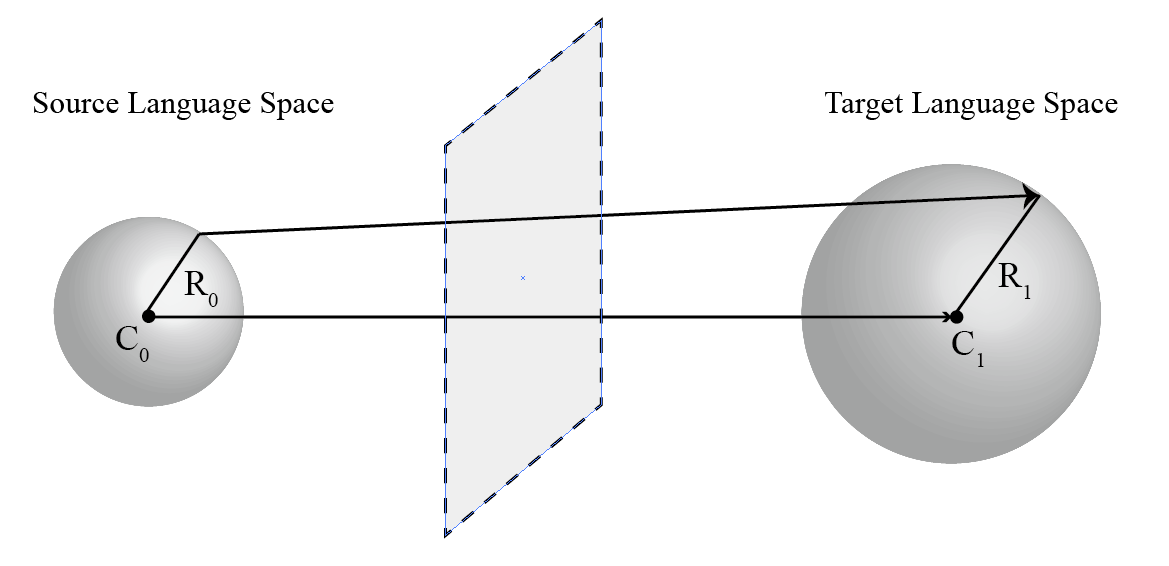}
		\caption{NE hypersphere transformation between two language spaces}
		\label{fig:hypersphere-trans}
	\end{figure}

	Instead of only discovering NE mapping, we try to find a general isomorphic mapping between embedding spaces at first and then apply it to map NE hypersphere from one to another language space. The isomorphic relations between embedding spaces exist due to the shared knowledge among different human languages. Meanwhile embedding learned for syntactic and semantic representation purpose is related to the shared knowledge. The main idea about building such a mapping is to minimize the total distance of the corresponding word pairs in different spaces. As each word vector in source space is transformed through linear transformation, all the transformation may be represented as a matrix. In this work, we adopt the Earth Mover's Distance (EMD) as the minimized objective \cite{cohen1999earth}. The resulted transformation matrix will be used to project the center vector and the radius of hypersphere model from the source embedding to the target one, as shown in Figure \ref{fig:hypersphere-trans}.
	
	The exact solution to minimize the distance between the source and the target embeddings has been proved to be NP-hard \cite{ding2017fptas}. However, a local optimal solution can be guaranteed by an alternating minimization process \cite{cohen1999earth}. Following the work of \cite{zhang2017earth}\footnote{\cite{artetxe2018acl} reported that the method used in \cite{zhang2017earth} works poorly on more realistic scenarios, and they proposed a new state-of-the-art unsupervised iterative self-learning solution for cross-lingual embedding mapping tasks. We tried their method but did not get better mapping results for our hypersphere transformation task.}, we adopt Wasserstein GAN \cite{arjovsky2017wasserstein} to form a transformation matrix, in which the generator $G$ transforms the source word embedding and intends to minimize the distance between the transformed source distribution and the target distribution. The critic $D$ measures the distance between the transformed source word embedding and the target word embedding, providing guidance for the generator $G$ during training process.
	
	To target a language on its NE detection, we suppose there is a source language with known NE hypersphere parameters by maximizing the detection performance according to a given NE dictionary, then the GAN learns a transformation matrix from two embedding spaces. The target NE hypersphere will be easily computed through the obtained transformation matrix.
	
	\section{Hypersphere for NE Recognition}
	
	As our NE hypersphere model may indicate the NE likelihood for any word with the distance from the hypersphere center, it potentially becomes a helpful cue in context dependent named entity recognition (NER). Thus we choose four strong NER models \cite{lample2016neural,Peters2018ELMO,akbik2018coling,shang2018learning} for the concerned NER task, all of which adopt BiLSTM-CRF \cite{huang2015bidirectional,lample2016neural} as backbone network structure, but with slight differences on the embedding or CRF layer\footnote{We choose these models due to their simplicity with remarkable performance so that we can focus on the effect of our hypersphere discovery.}. 
	
	In the following, we first outline the baseline taggers and then introduce our method on NE hypersphere enhancement.
	
	\begin{table*}
		\begin{minipage}[b]{.5\linewidth}
			\centering
			\begin{tabular}{lll}
				\toprule
				\bf Type & \bf Radius & \bf F1-score \\\hline
				Per & 0.914 & 0.177 \\
				Loc & 0.725 & 0.203 \\
				Org & 0.784 & 0.904 \\\hline
				Total & 0.816 & 0.243 \\\bottomrule
			\end{tabular}
			\centering
			\caption{ NE hypersphere on English }
			\label{f-en-64-d}
		\end{minipage}%
		\begin{minipage}[b]{.5\linewidth}
			\centering
			\begin{tabular}{lll}
				\toprule
				\bf Type & \bf Radius & \bf F1-score \\\hline
				Per & 0.631 & 0.176 \\
				Loc & 0.737 & 0.885 \\
				Org & 0.822 & 0.811 \\\hline
				Total & 0.746 & 0.826 \\\bottomrule
			\end{tabular}
			\caption{ NE hypersphere on Chinese }
			\label{f-zh-64-d}
		\end{minipage}
	\end{table*}
	\subsection{Baselines}
	\paragraph{BiLSTM-CRF}
	Following \cite{lample2016neural}, we employ bi-directional Long Short-Term Memory network (BiLSTM) for sequence modeling. In embedding layer, we concatenate the word-level and character-level embedding as the joint word representation. As word embeddings generalize poorly for rare and out-of-vocabulary words, we augment word embeddings with smaller units, i.e., characters. The character embedding\footnote{Character embeddings are used both in English and Chinese NER experiments.} is generated by taking the final outputs of a BiLSTM applied to the embeddings from a lookup table of characters. Characters $w = \{c_{1},c_{2},\dots,c_{l}\}$ of each word are successively fed to BiLSTM and the final hidden states from both directions form the character-derived word representation.
	
	The model performance for independent classification task is limited when faced with cases having strong connections or dependencies across output labels, though it shows high accuracy in simple tasks such as part-of-speech (POS) tagging. Especially in NE extraction tasks, the construction of a sequence contains various constraints which is potentially against the independence assumptions. Hence, rather than making tagging decisions independently, we integrate conditional random field (CRF) \cite{lafferty2001conditional} into our model. Concretely, we use LSTM for encoding process, followed by CRF for tagging decisions. 
	
	We define the input sequence $x = \left \{ {x_{1}},...,{x_{n}} \right \}$ where ${x_{i}}$ stands for the ${i}$th word in sequence $x$. And $y = \left \{ {y_{1}},...,{y_{n}} \right \}$ is a predicted sequence of tags for $x$. The probabilistic model for sequence CRF defines a family of conditional probability $p$. For all possible tag sequence $y$, with given $x$, $p$ can be defined as:
	\begin{equation}
	p\left ( y|x\right)=\frac{e^{s(x,y)}}{\sum_{\widetilde{y}\in y_{x}} e^{s(x,\widetilde{y})}}
	\end{equation}
	\newline in which $s(x,y)$ denotes the sum of scores of the tag sequence $y$ with given input $x$. Consider ${T_{i,j}}$ as the transition score from the $i$-th tag to the $j$-th tag and ${L_{i,j}}$ as the score of $j$-th tag for $i$-th word from BiLSTM. $s(x,y)$ can be described as:
	\begin{equation}
	s(x,y)=\sum_{i=0}^{n}T_{y_{i},y_{i+1}}+\sum_{i=1}^{n}L_{i,y_{i}}
	\end{equation}
	
	During the CRF training process, we consider the maximum log-likelihood of the correct NE tag sequence. $\left \{ ({x_{i}},{y_{i}}) \right \}$ denotes for the input of the training dataset. The training objective is to maximize the logarithm of the likelihood.
	
	Only considering the interactions between two successive tags, training and decoding can be solved efficiently for the sequence CRF model.
	
	\paragraph{Advanced baselines}
	Here we briefly outline the four advanced baseline models, all of which adopt the BiLSTM-CRF as the basic structure with modifications on the embedding used in the encoding layer.
	
	\cite{Peters2018ELMO} introduced ELMo to the input layer to give a concatenated word representation where ELMo was learned from the internal states of a deep bidirectional language model (biLM), which is pre-trained on 1B Word Benchmark \cite{chelba2013one}. 
	
	Similarly, \cite{akbik2018coling} leveraged the internal states of a trained character language model to produce a new type of word embedding, \emph{contextual string embeddings}, where the same word will have different embeddings depending on its contextual use. This embedding is then utilized in the BiLSTM-CRF sequence tagging module to replace the character embedding.

	Recently, \cite{devlin2018bert} proposed a new language representation model BERT (Bidirectional Encoder Representations from Transformers). Since BERT models tokens in subword-level, we feed each input word into the WordPiece tokenizer and use the hidden state of the first sub-token as the input to the downstream model.
	
	\cite{shang2018learning} proposed a distant supervised tagging scheme, \emph{Tie or Break} that focuses on the ties between adjacent tokens, i.e., whether they are tied in the same entity mentions or broken into two parts. Accordingly, their proposed \emph{AutoNER} is designed to distinguish \emph{Break} from \emph{Tie} while \emph{Unknown} positions will be skipped using BiLSTM. The output of the BiLSTM will be further re-aligned to form a new feature vector to fed into a softmax layer to estimate the entity type, without CRF layer and Viterbi decoding.

	\subsection{Hypersphere Enhancement}To use the hypersphere to guide neural models to discover the semantic features from spatial distributions, we calculate the $Euclidean$  as the same as Eq.(\ref{ED}) to measure the distance between each word $W$ and the hypersphere center $C$ of the three NE types. We then normalize the Euclidean distances using \emph{z-scores},
	\begin{equation}
        z = \frac{ED(W, C)-\mu }{\sigma }
	\end{equation}
	where $\mu$ is the mean of the population and $\sigma$ is the standard deviation. At last, the hypersphere feature is formed as 3-$d$ vector which is concatenated with word embeddings to feed NER model. 
	
	\section{Experiment}
	
	In this section, we evaluate the hypersphere model by conducting three kinds of NE experiments, \emph{monolingual hypersphere modeling}, \emph{cross-lingual hypersphere transformation} and \emph{NE recognition}. Our evaluation aims to answer the following empirical questions: 
	
	$\bullet$ Is NE hypersphere capable of sufficiently and accurately extracting NEs from monolingual word embeddings?
	
	$\bullet$ Can NE hypersphere perform effectively in cross-lingual tasks by adopting appropriate transformation?
	
	$\bullet$  Can NE hypersphere model enhance an existing context dependent NE recognition task?
	
	\subsection{Setup}
	We use pre-trained word embeddings on Wikipedia in 300-D and 64-D with Word2Vec skip-gram model \cite{mikolov2013distributed}. Word embeddings in 300-D are utilized in visualization and NE detection tasks, and 64-D word embeddings are used for monolingual hypersphere modeling and cross-lingual hypersphere mapping. For word segmentation, we use jieba\footnote{https://github.com/fxsjy/jieba} for Chinese and NLTK\footnote{https://github.com/nltk/nltk} for English.
	
	\begin{table*}
		\begin{minipage}[b]{.5\linewidth}
			\centering
			\begin{tabular}{p{0.55cm}p{1.8cm}p{2.4cm}p{1.4cm}}
				\toprule \bf Type &\bf $F_{en}(R_{en})$ &\bf $F_{en-zh}(R_{en-zh})$& F-Ratio \\ \hline
				Per & 0.177(0.914) & 0.072(0.972)& 0.405 \\
				Loc & 0.203(0.725) & 0.152(1.65)& 0.749 \\
				Org & 0.904(0.784) & 0.281(1.28)& 0.311 \\
				\hline
				Total & 0.243(0.816) & 0.138(1.53) & 0.568 \\
				\bottomrule
			\end{tabular}
			\centering
			\caption{ Cross-lingual Results of en-zh }
			\label{f-en-zh-64-d}
		\end{minipage}%
		\begin{minipage}[b]{.5\linewidth}
			\centering
			\begin{tabular}{p{0.55cm}p{1.8cm}p{2.4cm}p{1.4cm}}
				\toprule \bf Type &\bf $F_{zh}(R_{zh})$ &\bf $F_{zh-en}(R_{zh-en})$& F-Ratio \\ \hline
				Per & 0.176(0.631) & 0.032(0.768) & 0.183 \\
				Loc & 0.885(0.737) & 0.320(0.995) & 0.362 \\
				Org & 0.811(0.822) & 0.573(0.819) & 0.707 \\
				\hline
				Total & 0.826(0.746) & 0.224(1.05) & 0.271 \\
				\bottomrule
			\end{tabular}
			\caption{ Cross-lingual Results of zh-en }
			\label{f-zh-en-64-d}
		\end{minipage}
	\end{table*}
	\subsection{Monolingual Hypersphere}
	
	Given the NE dictionaries in Table \ref{table-sts-NE-dictionaries}, word embeddings in different dimensions occur to have different parameter values that maximize the NE detection F1-score, following the discussion in subsection 3.1. We pre-trained English and Chinese word embeddings in different dimensions for the purpose of finding a certain dimension that has the best performance. For each choice of the number of dimensions, we make a greedy parameter setting search to achieve such an aim. We found several local optima between the dimensions ranging from 16-D to 300-D\footnote{We choose the lower dimension for the following experiment when they have similar performances.}. At last, we found word embeddings in 64-D outperform the others both in English and Chinese. Hence, based on this conclusion, the following experiments are conducted on word embeddings in 64-D as well.
	
	Tables \ref{f-en-64-d} and \ref{f-zh-64-d} present the max F1-score of three types of NEs on English and Chinese along with the radius. 
	
	It is undeniable that there are differences among the three kinds of NEs, as well as between English and Chinese. Firstly, the quality of NE dictionaries exerts key influences on the results. Due to the incompleteness of NE dictionaries, the accuracy of determining the center vector and the radius can be undesirably effected and consequently cause influence on the final NE detection. Secondly, the performance of the proposed method is also largely influenced by the quality of word embeddings. Noises in pre-processing and deficient training data are possible factors that contribute to a decrease in performance.

	\subsection{Hypersphere Transformation}

	\begin{table*}
		\centering
		\begin{tabular}{lccccccccc}
			\toprule  
			& \multicolumn{3}{c}{\textbf{CoNLL 2003 (English)}} & \multicolumn{3}{c}{\textbf{CityU (Chinese)}} & \multicolumn{3}{c}{\textbf{MSRA (Chinese)}} \\
			& Sentence& Token& Entity& Sentence& Token& Entity& Sentence& Token& Entity\\
			\midrule
			Train& 15K& 204K& 23K & 44K& 2.41M & 101K& 40K& 1.96M & 68K\\
			Dev& 3.5K& 51K&5.9K & 4.8K & 294K& 11K & 4.4K& 213K & 7.5K\\
			Test& 3.7K& 46K& 5.6K& 6.3K& 364K&16K & 3.3K& 173K & 6.2K\\
			\bottomrule 
		\end{tabular}
		\caption{ Statistic of NE Datasets}
		\label{table-sts-ne-dataset}
		\centering
	\end{table*}
	Wasserstein GAN with the same setting as \cite{zhang2017earth} is used to minimize the selected Earth Mover's Distance between source and target word embedding in this task. By the learned transformation matrix, we can identify an NE range on the target language. The NE detection performance of the resulted hypersphere in the target language is also measured by F1-score as described in Section \ref{sec-mono-hyper}. We choose English and Chinese as the source and target languages respectively, based on the self-collected dictionaries shown in Table \ref{table-sts-NE-dictionaries} and Word2Vec word embedding trained on Wikipedia in 64-D. To individually evaluate the transformation performance without considering the monolingual modeling loss, we also report the ratio of F1-scores of transformed hypersphere and native best hypersphere.
	
	Our cross-lingual NE detection results are shown in Tables \ref{f-en-zh-64-d} and \ref{f-zh-en-64-d}. From the tables, we find that the transformed hypersphere models show satisfying results on cross-lingual mapping. Nevertheless, we observed that cross-lingual F1-score for Locations (Loc) in English is better than that in monolingual task. This may be attributed to the high quality of Chinese location NE dictionary so that transform matrix can additionally help the bilingual processing which right demonstrates the effectiveness of our approach. While for the monolingual case, we have too poor English location NE dictionary in the meantime. Similar to monolingual tasks, cross-lingual tasks are also effected by the quality of NE dictionaries and word embeddings. Actual F1-scores are expected to be higher than presented below due to the relatively poor quality of current dictionaries. To our best knowledge, there comes no previous work for cross-lingual NE embedding mapping task, so that we do not have previous similar systems for comparison purpose.
	
	\begin{table}
		\centering
		\resizebox{\linewidth}{!}
		{
			\begin{tabular}{lccc}
				
				\toprule  
				\multirow{2}{*}{\textbf{Model}}&  \multicolumn{2}{c}{\textbf{Baseline}} & \multirow{2}{*}{\textbf{Ours (ERR)}} \\
				& \textbf{Report}& \textbf{Our run} &\\
				\midrule
				\cite{lample2016neural} & 90.94 & 90.97 & 91.18 (2.3) \\
				\cite{Peters2018ELMO} & 92.22 & 92.58 & \textbf{93.10 (7.0)} \\
				\cite{akbik2018coling} & 93.09 & 92.72 &  92.95 (3.2)\\
				\cite{shang2018learning} & / & 84.68 & 85.31 (4.1)\\
				\cite{devlin2018bert} & 92.80 & 91.62 & 91.81 (2.6) \\
				\bottomrule 
			\end{tabular}
		}
		\caption{Results (\%) of English NER without and
			with hypersphese features. The baseline results are either from our re-implementation using the released codes or the report in the corresponding papers \protect\footnotemark. ERR in the brackets show the relative error rate reduction of our models to the baselines.}
		\label{table-ner-result-en}
		
		\centering
	\end{table}

	\begin{table}
		\centering
		{
			
			\begin{tabular}{lccc}
				\toprule  
				\textbf{Model}& \textbf{CityU} & \textbf{MSRA} \\
				\hline
				\cite{zhao2008unsupervised}& 89.18 & 86.30 \\
				\cite{zhou2013chinese} & 89.78& 90.28\\
				\cite{dong2016character} & /& 90.95\\
				\hline
				Baseline (BiLSTM+CRF) & 89.84& 89.93\\
				Ours (+hypersphere feature) & 90.24 & 90.98 \\
				Baseline (BERT) & 95.10 & 95.33\\
				Ours (+hypersphere feature) & \textbf{95.30} & \textbf{95.53} \\
				\bottomrule 
			\end{tabular}
		}
		\caption{Results (\%) of Chinese NER.}
		\label{table-ner-result-zh}
		\centering
	\end{table}
	
	\footnotetext{As researchers discussed in \url{https://github.com/zalandoresearch/flair/issues/206} and \url{https://github.com/google-research/bert/issues/223}, we could not reproduce those results with our best efforts, either. Here we only show it for reference, and we hope to focus on the improvements via our method.}

	\subsection{NE Recognition}
	
	To evaluate our model performance for NE recognition, we present the results on both English and Chinese datasets as shown in Table \ref{table-sts-ne-dataset}.

	The English dataset is from CoNLL 2003 shared task \cite{tjong2003introduction}, whose corpus includes four kinds of NEs: Person, Location, Organization and MISC. The Chinese datasets\footnote{Since training a Chinese ELMo or character based language model is quite time-consuming and AutoNER showed unsatisfactory performance for Chinese (around 40\% F1-score), we only report the results of BiLSTM-CRF baseline following the same architecture as that for English NER \cite{lample2016neural}.}, MSRA and CityU, are from Third SIGHAN Chinese Language Processing Bakeoff \footnote{Available at: http://sighan.cs.uchicago.edu/bakeoff2006/}. Table \ref{table-sts-ne-dataset} lists statistics of all datasets. We use \emph{OpenCC}\footnote{Available at: https://github.com/BYVoid/OpenCC} to transfer the traditional Chinese data into simplified text. Since there is no validation set for both of the Chinese datasets, we hold the last 1/10 for development within the training data. Statistics of CoNLL2003 and SIGHAN is shown in Table \ref{table-sts-ne-dataset}. Self-collected NE dictionaries listed in Table \ref{table-sts-NE-dictionaries} are used to calculated center vector and hypersphere distance feature. Hence, the quality of self-collected NE dictionaries also has a major impact on the performance of hypersphere enhancement. We adopt F1-score to evaluate the performance of models. We follow the same hyper-parameters for each model as the original settings from their corresponding literatures \cite{lample2016neural,Peters2018ELMO,akbik2018coling,shang2018learning} to isolate the impact of our hypersphere embeddings from earlier approaches.

	Our NER model is simply the baseline plus an NE hypersphere guide enhancement. The comparison is given in Tables \ref{table-ner-result-en} and \ref{table-ner-result-zh}, which shows that the hypersphere feature could essentially boost all the model performance substantially for both English and Chinese NER, achieving new state-of-the-art results. For Chinese evaluation, even though we use the same hyper-parameters as for English, our model also outperforms the BiLSTM-CRF baseline by a large margin, especially with 1.05\% improvement on MSRA dataset. 
	
	\section{Related Work}
	\label{sec:length}
	
	There are several NER works using the traditional annotation projection approaches \cite{yarowsky2001inducing,zitouni2008mention,mayhew2017cheap}. With parallel corpora, or translation techniques, they project NE tags across language pairs, such as \cite{pan2017cross}.\citeauthor{wang2013cross}~(\citeyear{wang2013cross}) proposed a variant of annotation projection which projects expectations of tags and uses them as constraints to train a model based on generalized expectation criteria. Besides this, annotation projection has also been applied to several other cross-lingual NLP tasks, such as word sense disambiguation in  \cite{diab2002unsupervised}, part-of-speech (POS) tagging in \cite{yarowsky2001inducing} and dependency parsing in \cite{rasooli2016cross}. Different from all the above works, we assume the inconvenient requirement about parallel or annotated corpora and start modeling from an intuitive visualization observation.
	
	For direct NE transformation, cross-lingual word clusters has been built by using monolingual data in source/target languages and aligned parallel data between source and target languages in \cite{tackstrom2012nudging}. The cross-lingual word clusters were then used to generate universal features. \citeauthor{tsai2016cross}~(\citeyear{tsai2016cross}) applied the cross-lingual wikifier developed in \cite{tsai2016cross} and multilingual Wikipedia dump to generate language-independent labels (FreeBase types and Wikipedia categories) for n-grams in \cite{tsai2016crossb}, and those labels were used as universal features. Different from the above methods, our model only relies on an embedding space aligning learning to project a simple bi-parameterized hypersphere. \citeauthor{collobert2011natural}~(\citeyear{collobert2011natural}) used gazetteer features for  substantial improvements.
	
	Most traditional sequence labeling models for NER are statistical models, including Hidden Markov Models (HMM) and Conditional Random Fields (CRF) \cite{passos2014lexicon,luo2015joint}, which rely heavily on hand-crafted features and task-specific resources. Such kind of context dependent NE recognition will heavily rely on a sufficient NE dictionary \cite{chen2006chinese} as a key knowledge driven feature. However, these resources are hard to obtain, especially for low-resource languages. Existing engineering methods count on manual collecting NE dictionaries which are vulnerable to insufficient NE collection and frequent updating. The proposed NE hypersphere gives a general mathematical model which can alleviate the NE resource difficulty, to some extent. Recent work also explored to use pre-trained language representations on large corpora \cite{Peters2018ELMO,devlin2018bert,akbik2018coling}. Compared with these models with high computational cost, our solution is much more lightweight while remaining effective.

	\section{Conclusion}
	
	Named entities are very hard to be well defined with a closed dictionary by considering that they keep emerging with new concepts and new knowledge through active language use. This paper presents a novel, open definition for monolingual NE definition, namely, a simple hypersphere model. We empirically show that NEs in different languages tend to gather into hypersphere through careful visualization on embedding space. The usefulness and the effectiveness of the proposed NE hypersphere model is verified in two NE recognition related applications. First, we propose using the Wasserstein GAN to learn a transformation matrix to map NE distribution between different languages, providing promising performance. Second, being a role of general NE dictionary, the NE hypersphere model guides the NE taggers over sentence to give significantly better performance. Without extra annotated data requirement on the given language besides the pre-trained embedding, we provide a potentially useful solution for effective NE detection with new state-of-the-art results.
	
	\bibliography{ner}
	\bibliographystyle{ner}

\end{document}